# REVIEW

# Let's Have a Chat! A Conversation with ChatGPT: Technology, Applications, and Limitations

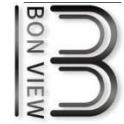

BON VIEW PUBLISHING

---

**Sakib Shahriar[1] and Kadhim Hayawi[2]**

1 School of Computer Science, University of Guelph, Canada.
2 College of Interdisciplinary Studies, Computational Systems, Zayed University, United Arab Emirates.

---

**Abstract:** The advent of AI-empowered chatbots capable of constructing human-like sentences and articulating cohesive essays has captivated global interest. This paper provides a historical perspective on chatbots, focusing on the technology underpinning the Chat Generative Pre-trained Transformer, better known as ChatGPT. We underscore the potential utility of ChatGPT across a multitude of fields, including healthcare, education, and research. To the best of our knowledge, this is the first review that not only highlights the applications of ChatGPT in multiple domains, but also analyzes its performance on examinations across various disciplines. Despite its promising capabilities, ChatGPT raises numerous ethical and privacy concerns that are meticulously explored in this paper. Acknowledging the current limitations of ChatGPT is crucial in understanding its potential for growth. We also ask ChatGPT to provide its point of view and present its responses to several questions we attempt to answer.

***Corresponding author**: Sakib Shahriar, School of Computer Science, University of Guelph, Canada. Email: shahrias@uoguelph.ca*

---

## 1. Introduction

In 1950, the British computer scientist Alan Turing disputed whether human reasoning can be matched by computers: "Can machines think?" (TURING, 1950). Subsequently, he proposed the Turing Test to measure computer or artificial intelligence. In a Turing test, a human interrogator is presented with responses from a human and a computer (with the ability to generate written texts in real-time). If the interrogator cannot distinguish between the answers, the computer system passes the Turing Test. Although several computer programs and chatbots like Eliza demonstrated success in the Turing test ((Weizenbaum, 1966) (Güzeldere & Franchi, 1995)), these programs arguably used certain tricks to pass the test (Pinar Saygin et al., 2000) rather than demonstrating any significant intelligence. With the advancement in machine learning and natural language processing (NLP), chatbots have gained significant research attention and have been used for a variety of commercial and non-commercial applications ((Luo et al., 2022), (Adamopoulou & Moussiades, 2020), (Ranoliya et al., 2017), (Rahman et al., 2017), (Zhou et al., 2020)). Despite their vast adoption, most chatbots do not have personalization, and user satisfaction remains questionable (Følstad & Brandtzaeg, 2020). This limitation prompted researchers and developers to focus on chatbot engagement in making chatbots more conversational.

In November 2022, OpenAI released a novel AI-powered chatbot called ChatGPT[1], which rapidly gained substantial public attention. Distinguishing itself from prior chatbots, ChatGPT has demonstrated notable proficiency in generating logically consistent responses to a broad array of inquiries. In this review, we investigate the technological evolution that facilitate ChatGPT's remarkable capabilities and illuminate various domains where this innovative tool finds its applications. Moreover, we leverage the conversational capabilities of ChatGPT to elicit its responses to a series of queries we sought to address.

---

[1] https://openai.com/blog/chatgpt/

> **Question:** *Hi ChatGPT, tell us about the history of chatbots*
>
> **ChatGPT:** *Certainly! The idea of a machine that could simulate human conversation dates back several decades. In fact, the first chatbot was created in the 1960s by MIT computer scientist Joseph Weizenbaum. Weizenbaum's chatbot, called ELIZA, used a set of rules to simulate conversation by asking questions and giving scripted responses. ELIZA was designed to imitate a psychotherapist, and people were often surprised at how convincingly it could simulate human conversation.*
>
> *Over the years, chatbots have continued to evolve and improve. One of the key advancements was the development of natural language processing (NLP) technology, which allows chatbots to understand and respond to human language more accurately. In the 1990s and early 2000s, chatbots began to be used for customer service and support, with companies like AOL and Microsoft developing chatbots to help users with their software and services.*
>
> *In recent years, there has been a surge of interest in chatbots, driven in part by advancements in artificial intelligence and machine learning. Modern chatbots use a combination of NLP, machine learning, and other AI technologies to understand and respond to human language in a more natural and nuanced way. They are now used in a wide range of applications, from customer service and support to personal assistants and entertainment.*
>
> *Overall, the history of chatbots is a fascinating one that reflects the evolution of AI and our growing desire to create machines that can understand and interact with us in more human-like ways.*

Our initial analysis reveals that the responses generated by ChatGPT exhibit commendable coherence, engagement, and grammatical precision. Furthermore, we observe that ChatGPT performs exceptionally well in disseminating factual content, demonstrating an impressive accuracy in specifying names and dates. Interestingly, ChatGPT refrains from providing information about itself, and it is evident that its knowledge is bounded, extending only up to September 2021[2]. Lastly, it is worth noting that ChatGPT adeptly concludes its responses with a summary that encapsulates its perspective, thus adding an engaging element to its communication.

## 2. ChatGPT Technology: From Language Models to Reinforcement Learning

Machine learning, a subfield of artificial intelligence (AI), is defined as the ability of computer systems to learn from experience without being explicitly taught or programmed. With advances in computing power, increasing data availability, and algorithmic improvements, deep learning has emerged as a high-performing predictive tool. Unlike traditional machine learning, deep learning algorithms use artificial neural networks, often in a large capacity. Deep learning models generally require more training data but can provide better generalization. Transformers or large language models are a special family of deep learning models that have emerged as state-of-the-art solutions for text data. From text classification to text generation, these models offer superior performance than other AI models. Language models are usually pre-trained on large language corpus, such as Wikipedia, but can be fine-tuned for a specific task. The two most prominent language models include Bidirectional Encoder Representations from Transformers (BERT) (Devlin et al., 2019) and Generative Pretrained Transformer (GPT) (Radford et al., 2019). The former is primarily used for recognizing the context in language and classification tasks, while the latter is more adept at language generation tasks such as machine translation and question-answering.

Several chatbots prior to ChatGPT used language models. However, the main component that allowed ChatGPT its coherent and engaging responses is attributed to the use of reinforcement learning. In this context, human evaluators ranked the responses of ChatGPT, which allowed it to optimize some of its parameter to essentially become a better conversationalist. ChatGPT was developed on top of GPT-3, a generative autoregressive language model with 175 billion parameters (Brown et al., 2020). To train such a complex network, OpenAI partnered with Microsoft with a reported training resource of over 350GB of memory costing $12 million[3]. More precisely, ChatGPT was built on GPT 3.5, which belongs to the family of large language models (LLMs). LLMs are advanced neural-network-based systems that leverage extensive model architectures and vast training datasets to assign probabilities to sequences of words, often exceeding 100 billion parameters (Carlini et al., 2021). These models undergo training on expansive datasets, enabling them to discern statistical patterns and associations within sequences of words. Consequently, they acquire the ability to generate or interpret language with remarkable proficiency. In these models, both input and output are

---

[2] https://www.forbes.com/sites/ariannajohnson/2022/12/07/heres-what-to-know-about-openais-chatgpt-what-its-disrupting-and-how-to-use-it/?sh=41280f462643

[3] https://venturebeat.com/ai/ai-machine-learning-openai-gpt-3-size-isnt-everything/

organized in the form of tokens, which are numerical representations of words. This numerical representation facilitates efficient processing by the model. The training process for the GPT model involves predicting the subsequent token given a sequence of input tokens, thus it essentially learns the language's structure. As a result, the model has the capability to generate text that is not only grammatically consistent but also semantically analogous to the data it was trained on. Once the model has been trained, it can be directed to perform specific language tasks like answering questions. This is achieved through the use of meticulously designed instructions or prompts, a process referred to as 'prompt engineering' (Liu & Chilton, 2022). However, ensuring that the model's output aligns with human values and ethics is of utmost importance to prevent the model from generating responses that could be harmful or unethical. To achieve this alignment, a technique called Reinforcement Learning from Human Feedback (RLHF) is employed. This process involves training the model to favor outputs that human evaluators rate as preferable, helping to guide the model's responses in a more ethical and useful direction.

---

**Question:** *Hi ChatGPT, how were you were developed?*
**ChatGPT:** *Hello! I was developed by OpenAI, a research organization focused on advancing artificial intelligence in a safe and beneficial way. Specifically, I was developed by a team of researchers and engineers at OpenAI who used a deep learning technique called a transformer to train me on a large corpus of text data. The transformer is a type of neural network that is especially good at processing sequential data, like language.*
*To train me, the team fed me huge amounts of text from a variety of sources, including books, articles, and websites. They then used a process called unsupervised learning to teach me to recognize patterns in the text data. Through this process, I learned to understand and generate natural language.*
*After my initial development, I went through a rigorous testing and refinement process to ensure that I was accurate, reliable, and safe to use. And since my release, I have been continually updated and improved to better serve my users.*

---

Similar to the initial response, the subsequent communication from ChatGPT continued to impress us with its engaging demeanor and well-structured format. We noted that ChatGPT accurately identifies the utilization of unsupervised learning for pattern recognition, an assertion that aligns with the descriptions found in the GPT paper by (Radford et al., 2019). Additionally, a noteworthy observation is that while describing the final phase, ChatGPT refers to it as a 'testing and refinement process,' without alluding to the specifics of reinforcement learning or any other detailed facets of the process.

## 3. Applications of ChatGPT

Despite being released to the public domain very recently, ChatGPT has already gained significant research attention. In this section, we describe recent works investigating the use of ChatGPT for various research and applications. This paper employs a scoping review approach to explore recent research focusing on the application of ChatGPT across various domains. This type of review is particularly suitable in this context due to the rapid pace of AI advancements. This review style allows us to capture a broad spectrum of research on the topic, providing insights on emerging evidence and identifying gaps in the current knowledge base. To conduct a comprehensive literature search, we utilized Google Scholar as our primary search engine. Google Scholar was chosen because of its expansive index of scholarly literature across various disciplines and formats. Our search included a combination of terms such as "ChatGPT" and "Large Language Models (LLMs)" to ensure the inclusion of all potentially relevant works. It is important to note that some of the papers included in this review were not peer-reviewed at the time of writing. We chose to include these articles due to the fast-paced nature of the AI field, where new advancements and findings may not yet have undergone the often lengthy peer-review process. The inclusion of such articles offers a more current and comprehensive overview of the field.

Thorp (Thorp, 2023) provided a qualitative assessment of ChatGPT responses to research topics, such as education, literature, and scientific writing. ChatGPT provided an entertaining writeup when asked to complete a scene from a classic play. In terms of education, ChatGPT can provide factual answers but still has some way to go in writing essays. In another study (Else, 2023), researchers were asked to distinguish between abstracts of scientific papers written by ChatGPT and humans. The results are surprising as human evaluators only identified 68% of the abstracts to be generated by ChatGPT. De Angelis et al. (De Angelis et al., 2023) discussed the evaluation of language models in light of ChatGPT and highlighted potential ethical and practical challenges in medicine and public health. The main challenges include the potential of AI-driven misinformation or "infodemic" that is often difficult to discern.

In the field of medicine and public health, ChatGPT has already been explored for various applications. Khan et al. (Khan et al., 2023) discussed several potential applications of ChatGPT in medical education, including personalized learning and generating case studies. The authors also pointed out that ChatGPT can be used in clinical management for documentation and decision support. Rao et al. (Rao et al., 2023) evaluated the effectiveness of ChatGPT in providing clinical decision support in radiology. The authors provided ChatGPT with text prompts such as "For variant 'Breast cancer screening. Average-risk women: women with <15% lifetime risk of breast cancer.', determine the single most appropriate imaging procedure" to evaluate its efficacy in breast cancer screening and breast pain detection. ChatGPT performed relatively well for the former task with 88.9% correct responses but only managed 58.3% correct responses for breast pain. The role of ChatGPT and generative AI in helping urologists has also been discussed (Gabrielson et al., 2023). ChatGPT can primarily help urologists in low-complexity tasks, giving them more time to focus on patients. Hulman et al. (Hulman et al., 2023) utilized ChatGPT to answer frequently asked questions about diabetes and asked healthcare employees to distinguish between human and ChatGPT-generated answers. The authors found that the evaluators could identify answers generated by ChatGPT correctly 59.5% of the time. The authors also concluded that despite ChatGPT not being trained exclusively in medical data, it has clinical knowledge and can identify information about disease management. Generating a medical report about a given topic may be useful in pharmaceutical education. To this end, Zhu et al. (Zhu et al., 2023) prompted ChatGPT to generate a mini-review on "lipid-based drug delivery systems." The authors concluded that ChatGPT can structure the topic well with meaningful conclusions for the readers. However, there are question marks over the accuracy due to a lack of reliable citations. Shen et al. (Shen et al., 2023) summarized other potential use cases and implications for ChatGPT in medicine.

Researchers also investigated whether ChatGPT can answer medical exam questions. Kung et al. (Kung et al., 2023) tested the performance of ChatGPT on the US medical licensing exam, consisting of three standardized tests required for medical licensure in the US. ChatGPT performed at the passing threshold level with 60% accuracy without specialized input from humans. Any questions containing visual information, such as medial images, were removed. The results demonstrate the potential of ChatGPT for medical education and assistance in clinical-decision making. However, in a Chinese national medical licensing exam, ChatGPT performed considerably lower, with 45.8% correct answers (Wang et al., 2023). In Ophthalmology, ChatGPT was tested with questions from the Ophthalmic knowledge assessment program containing two exams and obtained 55.8% and 42.7% accuracy, respectively (Antaki et al., 2023). For a basic and advanced cardiovascular life support exam from the American Heart Association, ChatGPT performed below the 84% passing threshold (Fijačko et al., 2023). However, its ability to provide detailed rationality with reasonable accuracy makes it a potentially useful tool for self-learning and exam preparation. Mbakwe et al. (Mbakwe et al., 2023) argued that the success of ChatGPT in answering medical examinations boils down to the nature of these exams being rote memorization rather than testing analysis and critical thinking.

One of the significantly anticipated applications of chatbots is in the domain of education. AI and technology can be effective in education in several aspects, including personalized learning (Shahriar et al., 2022). In this context, ChatGPT can enhance student participation, provide experiential learning, and help educators in the evaluation of exams and content preparation (Kasneci et al., 2023). Several researchers focused their studies on the impact of ChatGPT in education ((Rudolph et al., 2023), (Lund & Wang, 2023), (Mhlanga, 2023), (Kasneci et al., 2023)). Potential concerns of ChatGPT in education include response bias ((Lund & Wang, 2023) (Mhlanga, 2023)), cheating (Rudolph et al., 2023), leakage of private data ((Lund & Wang, 2023) (Mhlanga, 2023)), and transparency (Mhlanga, 2023)). Chatbots can also contribute effectively to peer tutoring. Pardos and Bhandari (Pardos & Bhandari, 2023) found that 70% of the hints offered by ChatGPT in elementary and intermediate Algebra topics could result in positive learning gains for students. Frieder et al. (Frieder et al., 2023) evaluated the mathematical capabilities of ChatGPT in helping Mathematicians with tasks like question answering and finding theorems. The researchers found that ChatGPT displayed a level of Mathematics proficiency below those of an average graduate student. There is public concern about ChatGPT being used for plagiarism, and it is necessary to create tools to detect such plagiarism. To this end, Khalil and Er (Khalil & Er, 2023) attempted to utilize ChatGPT as a plagiarism detection tool and found that responses created by ChatGPT can often go undetected by other plagiarism checkers. Meanwhile, when ChatGPT was asked to check if the writing was generated by itself, it performed better than other plagiarism detection software. Yang et al. (Yang et al., 2023) explored ChatGPT's ability in summarizing written texts and found that ChatGPT performs on par with existing fine-tuning methods based on Rouge scores. The authors also highlighted that the current maximum input token length of 5000 is a limitation in assessing ChatGPT's ability in text summarizing. Can ChatGPT improve student essays? To answer this question, Basic et al. (Basic et al., 2023) conducted a study with nine students in the control group and nine in the experimental group that used ChatGPT. The authors concluded that ChatGPT does not necessarily improve essay quality because the control group outperformed the experimental group in most criteria.

Bang et al. (Bang et al., 2023) evaluated the effectiveness of ChatGPT on ten reasoning tasks, such as logical reasoning and commonsense reasoning. The authors found that ChatGPT performs reasonably well on deductive reasoning tasks but extremely poor on inductive reasoning. ChatGPT also does well on commonsense reasoning. For instance, ChatGPT was asked the following question; *Please judge if this predicate is (likely) plausible or implausible: "knife chop rock."* To which it answered: *It is implausible that a knife could chop through a rock. Knives are made to cut through softer materials such as food, paper, or wood. Rocks are much harder than these materials and a knife would not be able to make a dent in a rock, let alone chop it.*

ChatGPT remains a potentially beneficial tool in journalism and misinformation detection, particularly for references and fact-checking. During the COVID-19 pandemic, vaccine misinformation led to vaccine hesitancy in social media (Hayawi et al., 2022). Sallam et al. (Sallam et al., 2023) provide a descriptive analysis of ChatGPT's response to vaccine conspiracy statements and compulsory vaccination. The study found that ChatGPT was dismissive of conspiracy statements but provided divided opinions for vaccine mandates by highlighting  the pros and cons.

In software development, ChatGPT can be used for tutoring and software testing. Jalil et al. (Jalil et al., 2023) evaluated ChatGPT's capabilities in software testing by presenting it with textbook questions. The results are not promising in this case as ChatGPT managed to answer correctly only 55.6% of the answers. Jiao et al. (Jiao et al., 2023) evaluated ChatGPT's performance in translation on several benchmark sets. In high-resource European languages, ChatGPT performs on par with commercial applications like Google Translate. However, for low-resource or distant languages, the performance is significantly worse. At this point, let us ask ChatGPT about its potential use in various domains.

---

**Question:** *Hi ChatGPT, in what ways can you be used in healthcare, education, and research?*

**ChatGPT:** *Hi! As an AI language model, I can be used in various ways in healthcare, education, and research. Here are some examples:*

*In Healthcare:*
*1) Medical Diagnosis and Treatment: AI language models can be trained to analyze medical data such as medical images, electronic health records, and other patient information to assist in medical diagnosis and treatment. This can help healthcare professionals make more accurate and informed decisions regarding patient care.*
*2) Patient Interaction: AI language models can be used to help patients interact with healthcare providers and provide them with information about their health conditions, treatment options, and medication instructions.*
*3) Medical Research: AI language models can help researchers process and analyze large volumes of medical data and identify patterns and trends in disease outbreaks, drug efficacy, and patient outcomes.*

*In Education:*
*1) Language Learning: AI language models can help language learners improve their language proficiency by providing them with personalized feedback, practice exercises, and conversation simulations.*
*2) Educational Content Creation: AI language models can be used to generate educational content such as quizzes, flashcards, and study guides that are customized to meet the needs of individual learners.*
*3) Academic Research: AI language models can be used to analyze academic texts and generate summaries, insights, and recommendations that can help researchers gain new insights and discoveries.*

*In Research:*
*1) Data Analysis: AI language models can be used to analyze large volumes of data from various sources and identify patterns, trends, and correlations that are not easily visible to human researchers.*
*2) Text Mining: AI language models can help researchers extract relevant information from large volumes of unstructured text data such as research papers, news articles, and social media posts.*
*3) Research Collaboration: AI language models can help researchers collaborate with each other by providing real-time translation services, automated proofreading and editing, and content summarization.*

**Question:** *What about journalism and software development?*
**ChatGPT:** *Certainly! As an AI language model, I can also be used in journalism and software development in several ways:*

*In Journalism:*
*1) News Writing and Summarization: AI language models can be used to generate news articles and summaries automatically, saving time and resources for journalists and news organizations.*
*2) Fact-Checking: AI language models can help fact-check news stories and verify the accuracy of the information presented.*
*3) Personalization: AI language models can help personalize news content for individual readers based on their interests, reading history, and location.*

*In Software Development:*
*1) Chatbots: AI language models can be used to power chatbots, which can interact with users and answer their questions in natural language.*
*2) Code Generation: AI language models can be used to generate code automatically, reducing the time and effort required for software development.*
*3) Bug Detection: AI language models can be used to detect bugs and errors in code, improving the quality and reliability of software applications.*

**Question:** *In what other domains can you be used for?*
**ChatGPT:** *As an AI language model, I can be used in a wide range of domains beyond healthcare, education, research, journalism, and software development. Here are a few examples:*

*1) Customer Service: AI language models can be used to power chatbots and virtual assistants that can provide customer service and support to users.*
*2) Financial Services: AI language models can be used to analyze financial data and detect fraud, as well as provide personalized financial advice to individuals.*
*3) Marketing and Advertising: AI language models can help marketers generate ad copy and content, analyze customer sentiment and behavior, and personalize marketing campaigns for individual users.*
*4) Legal Services: AI language models can be used to analyze legal documents and provide legal advice to individuals and businesses.*
*5) Social Media: AI language models can be used to analyze social media content, monitor online conversations, and detect patterns and trends in user behavior.*

*6) Gaming: AI language models can be used to generate dialogue and narrative content for video games, as well as power virtual assistants and chatbots within games.*
*7) Transportation: AI language models can be used to power voice assistants in vehicles and help drivers navigate and control various vehicle systems.*
*8) Robotics: AI language models can be used to enable natural language interaction between humans and robots, as well as help robots understand and respond to human commands and requests.*
*These are just a few examples of how AI language models like me can be used in various domains. The possibilities are endless, and new applications are being developed every day as AI technology advances.*

ChatGPT has raised a lot of eyebrows for its potential use in writing scientific articles. According to Stokel-Walker (Stokel-Walker, 2023), four papers have listed ChatGPT as a co-author for its writing contributions. However, leading publishers like Science have dismissed the possibility of listing an AI chatbot as a co-author while other publishers, including Taylor & Francis, are reviewing their policy when it comes to this issue. Several journals, including the Korean Journal of Radiology, have published policies rejecting any contributions as co-authors from ChatGPT or other AI models (Park, 2023). On the other hand, other journals published guidelines recommending the acknowledgment of any AI tools used for the research but do not allow them to be listed as co-authors ((Thornton et al., 2023), (Polesie & Lark ú, 2023)). Researchers have also called on journals to clarify what proportion of their papers contain AI-generated content (Tang, 2023) and publish guidelines for AI use in writing papers (Aczel & Wagenmakers, 2023). Korinek (Korinek, 2023) explored the potential use cases of language models like ChatGPT for Economic research. The author argues that researchers can be more productive by using language models for tasks like editing and generating headlines. Chen (Chen, n.d.) discussed some of the ethical concerns and potential benefits of using AI tools for scientific writing. The author argued that chatbots can assist writers whose native language is not English. The paper was written by the author in Chinese and summarized by ChatGPT and translated to English by AI tools. Aydın and Karaarslan (Aydın & Karaarslan, 2022) utilized ChatGPT to write a literature review on the role of Digital twins in healthcare. Despite the promising results, the authors found that ChatGPT had significant matches on a plagiarism checker when paraphrasing sentences. Dowling and Lucey (Dowling & Lucey, 2023) found that ChatGPT is effective in generating plausible research ideas, literature reviews, and testing frameworks. They also noted that the research quality improves significantly when domain expertise is added as input. Although ChatGPT can potentially speed up research and writing of scientific papers, there should be human oversight and fact-checking as language models like ChatGPT may generate misleading information ((Stokel-Walker & Van Noorden, 2023), (Lee, 2023), (Lin, 2023), (Alkaissi & McFarlane, 2023)). Table 1 summarizes the existing works utilizing ChatGPT in several domains.

**Table 1 Applications of ChatGPT in Existing Works**

| Domain | Application | Method | Results |
|---|---|---|---|
| **Medicine and Public Health** | Clinical decision support in radiology (Rao et al., 2023) | Provided text prompts and evaluated diagnosis response for breast cancer screening | Effective in breast cancer screening (88.9%) but not in breast pain detection (58.3%) |
| | Improving efficiency in Urology (Gabrielson et al., 2023) | Discussed the role of generative AI in helping urologists | N/A |
| | Answer frequently asked medical questions (Hulman et al., 2023) | Asked human evaluators to distinguish between ChatGPT and human answers to diabetes-related questions | The evaluators could only identify answers to be generated by ChatGPT 59.5% of the time |
| | Write a short review on a medical topic (Zhu et al., 2023) | ChatGPT was asked to write a mini review on lipid-based drug delivery systems | ChatGPT can benefit readers with general knowledge but the accuracy of analysis is questionable due a lack of reliable citations |
| | Take on medical exam (Kung et al., 2023) | Asked ChatGPT questions from US medical licensing exam after removing questions with visual information | ChatGPT performed near the passing threshold of 60% accuracy |
| | Take on a medical exam in Chinese (Wang et al., 2023) | Asked ChatGPT questions from the Chinese national medical licensing exam | ChatGPT performed considerably lower, with 45.8% correct answers |
| | Take on Ophthalmology exam (Antaki et al., 2023) | Asked ChatGPT questions from Ophthalmic knowledge assessment program | Obtained 55.8% and 42.7% accuracy on the two exams, respectively |
| | Take on a life support exam (Fijačko et al., 2023) | Asked ChatGPT questions from basic and advanced cardiovascular life support exams by the American Heart Association | Despite scoring below the passing threshold (84%), ChatGPT showed promising results by providing detailed explanation of its rationale |
| **Education** | Analyze impact of ChatGPT in education ((Rudolph et al., 2023), (Lund & Wang, 2023), (Mhlanga, 2023)) | Highlighted potential benefits and ethical concerns | N/A |
| | Algebra tutoring (Pardos & Bhandari, 2023) | Used ChatGPT to generate hints for Algebra topics | 70% of the hints could result in positive learning gains for students |
| | Assess Mathematics proficiency of ChatGPT (Frieder et al., 2023) | Created a dataset containing questions related to elementary arithmetic problems, symbolic problems, and other exercises | For most problems, ChatGPT does not cross the passing threshold (50%) |
| | Plagiarism Detection Tool (Khalil & Er, 2023) | Asked ChatGPT to determine if a writing was generated by itself | Performed better than other plagiarism detection software with 92% accuracy |
| | Text Summarization (Yang et al., 2023) | Compared ChatGPT's performance in summarizing texts with other fine-tuning models | Similar performance to existing models in terms of Rouge scores (30.94 Average R-1) |
| | Improve essay quality (Basic et al., 2023) | Divided students into control and experimental groups (who used ChatGPT) to assess their essays | No evidence of ChatGPT improving writing quality as control group outperformed experimental group |
| **Reasoning** | Evaluate ChatGPT on various reasoning tasks (Bang et al., 2023) | Proposed framework to evaluate the multitask, multilingual, and multimodal reasoning of language models | Performs well on deductive and commonsense reasoning but not on inductive reasoning. Overall, 63.41% average accuracy in 10 different reasoning categories |
| **Journalism and Misinformation Detection** | Evaluate ChatGPT response on conspiracy statements and politically divided ideas (Sallam et al., 2023) | Performed a descriptive study on ChatGPT responses to various COVID-19 vaccine topics | ChatGPT was dismissive of conspiracy statements but remained neutral on controversial political views by stating pros and cons |
| **Software Development** | Evaluate ChatGPT's response to software testing questions (Jalil et al., 2023) | Asked ChatGPT to respond to textbook questions in a common software testing curriculum | Only 55.6% of the questions were correctly answered |
| **Translation** | Evaluate ChatGPT's ability in translation (Jiao et al., 2023) | Used several benchmark test sets to compare ChatGPT's responses in translating | Performs on par with commercial applications for high-resource European languages but lags behind on distant languages |

| Scientific Research | Write scientific and academic papers (Stokel-Walker, 2023) | Discussed several papers who used and listed ChatGPT as authors | Highlighted potential ethical concerns |
|---|---|---|---|
| | Automation in Economic research (Korinek, 2023) | Highlighted use cases of ChatGPT in research, including language editing and headline generation | N/A |
| | Summarize and translate research papers for non-native English speakers (Chen, n.d.) | Used ChatGPT to summarize and other AI tools to translate Chinese written paper | Coherent writing style after manual screening. Potential benefit in speeding up research but has several ethical concerns |

## 4. Analysis of ChatGPT Performance on Exams

In the previous section, we discussed some of ChatGPT's performance on taking on medical exams. ChatGPT's effective exam performance can facilitate personalized learning, exam preparation, and tutoring services. In addition, it can also provide instantaneous feedback to students and create a supportive learning environment for students, resulting in an effective and convenient tool for students to prepare for exams and improve their scores.

In this section we summarize the performance of ChatGPT on different exams and present an overall aggregate. It must be noted that some work did not present quantitative results. Table 2 highlights the performance of ChatGPT on several exams. Overall, ChatGPT obtained an average accuracy of 59.53% across all exams.

An initial observation from this data is the broad range in accuracy results, indicating that ChatGPT's performance is highly dependent on the specific exam or domain in question. In terms of high performance, it is noticeable that ChatGPT scored remarkably well in the United States Medical Licensing Exam, with accuracies of 89.5% and 94.6% reported by (Fijačko et al., 2023) and (Kung et al., 2023), respectively. This suggests that ChatGPT can be highly effective in understanding and answering complex medical-related questions. On the other hand, ChatGPT struggled with certain subjects. Notably, (Antaki et al., 2023) reported relatively low accuracies of 55.8% and 42.7% in two Ophthalmology exams. Similarly, (Wang et al., 2023) reported low performance for the Chinese National Medical Licensing Examination, raising questions about the model's adaptability and consistency in certain subject areas.

Interestingly, (Choi et al., 2023) presented a dichotomy within the same domain of Law Examination, with Constitutional Law yielding 84.0% accuracy, while Taxation, particularly Mathematical questions, demonstrated the lowest accuracy in the table at 27.6%. This implies that the model may have difficulty handling math-related problems or topics that require numerical reasoning. Two studies, (Susnjak, 2022) and (Uludag, 2023) reported qualitative results for various subjects and Psychology, respectively. The absence of numerical performance indicators makes it challenging to directly compare these results with others in the table, emphasizing the need for more uniform evaluation metrics. Finally, the average accuracy across these reported studies is approximately 59.53%, which points to a generally competent performance, but with substantial room for improvement, particularly in domains where the accuracy was significantly lower. The performance variations suggest that further refinements to ChatGPT's training and fine-tuning processes might be required to enhance its proficiency across diverse fields.

### Table 2 ChatGPT Performance on Various Exams

| Reference | Exam | Performance (Accuracy) |
|---|---|---|
| (Fijačko et al., 2023) | United States Medical Licensing Exam | 89.5% |
| (Antaki et al., 2023) | Ophthalmology (Exam 1) | 55.8% |
| | Ophthalmology (Exam 2) | 42.7% |
| (Wang et al., 2023) | Chinese National Medical Licensing Examination (2021) | 45.8% |
| | Chinese National Medical Licensing Examination (2022) | 36.5% |
| (Kung et al., 2023) | United States Medical Licensing Exam | 94.6% |
| (Susnjak, 2022) | Education, Machine Learning, History, and Marketing. | N/A (Reported results are qualitative) |
| (de Winter, 2023) | High school exams on English Language Comprehension | 79.8% |
| | Law Examination: | 84.0% |

| (Choi et al., 2023) | Constitutional Law | |
|---|---|---|
| | Torts | 60.0% |
| | Taxation (Mathematical questions) | 27.6% |
| | Taxation (non-Mathematical questions) | 51.6% |
| (Uludag, 2023) | Psychology | N/A (Reported results are qualitative) |
| (Bordt & von Luxburg, 2023) | Computer Science Exam | 60.0% |
| (Jalil et al., 2023) | Software Testing Education | 49.4% |
| (Gilson et al., 2022) | Medical Licensing Examination: AMBOSS-Step1 | 44.0% |
| | AMBOSS-Step2 | 42.0% |
| | NBME-Free-Step1 | 64.4% |
| | NBME-Free-Step2 | 57.8% |
| (AL-Qadri & Ahmed, 2023) | Principles of Statistics Exam | 93.0% |
| Average | | 59.53% |

## 5. Limitations of ChatGPT

ChatGPT certainly has the potential for diverse and interesting applications. However, users should consider the limitations of the current model. In this section, we outline some of the current limitations of ChatGPT.

ChatGPT may sound interesting and convincing, but don't take its word for it! Indeed, ChatGPT's ability in forming meaningful and conversational sentences is quite impressive, but it may often 'hallucinate' responses (Alkaissi & McFarlane, 2023). Therefore, it is strongly recommended to verify and fact-check any responses from ChatGPT.

ChatGPT makes errors in simple reasoning, logic, mathematics, and presenting factual information (Borji, 2023). It is likely that the next version of GPT-4, expected to be released sometime in 2023, will significantly improve ChatGPT. According to several sources, GPT-4 network will be far more complex than its predecessor containing around 100 trillion parameters[4]. The GPT-3 model, in comparison, is made up of 175 billion parameters.

ChatGPT is currently limited in processing a maximum of 5000 tokens of text as input. While this is not a problem in most applications, it can be challenging in tasks like text summarization. Moreover, the current interface of ChatGPT does not allow uploading images or audio files. ChatGPT can produce code representations of visual images based on text prompts, but its drawing skills are somewhat limited currently (Bang et al., 2023). In this context, researchers have recently introduced a multi-modal language model trained on multi-modal corpora like image-caption pairs and can perceive general modalities (Huang et al., 2023). Multi-modal systems can provide applications, such as image generation from text prompts and stem isolation from pop music[5].

ChatGPT has a tendency to give wordy and detailed responses unless explicitly asked not to. Moreover, ChatGPT expresses fewer emotions than an average human and tends to be more objective (Guo et al., 2023). Therefore, ChatGPT cannot replace the need for human connection or be your friend! Similarly, it cannot be used for personal therapy or counseling, which require an intimate human connection.

Although ChatGPT can fetch you information about an incident prior to September 2021, do not expect it to give you the latest news! When asked about the recent Earthquake in Turkey, ChatGPT replies: '*I'm sorry, but as an AI language model, I do not have access to information from the future. As of my knowledge cutoff date of September 2021, there had not been any significant earthquakes in Turkey in 2023. However, earthquakes can occur at any time, and it's important to be prepared and informed about potential risks and safety measures.*'

---

**Question:** *Hi ChatGPT, what are some of your limitations?*
**ChatGPT:** *As an AI language model, I have some limitations, some of which include:*

---

[4] https://www.forbes.com/sites/bernardmarr/2023/02/24/gpt-4-is-coming--what-we-know-so-far/?sh=206f0ae86c2d

[5] https://aianswerz.com/

*1) Limited context: While I can understand and respond to a wide range of topics and questions, my responses are limited to the information and context provided in the question or statement. I do not have the ability to understand context beyond the given information.*
*2) Lack of emotion and empathy: I do not have emotions, and I cannot empathize with the user or understand their feelings.*
*3) Inability to interpret nonverbal cues: As an AI language model, I do not have the ability to interpret nonverbal cues, such as body language or facial expressions, which are essential in human communication.*
*4) Dependence on data quality: My responses are only as good as the data used to train me. If the data used to train me is biased or incomplete, my responses may also be biased or incomplete.*
*5) Inability to perform physical tasks: I am a language model and do not have a physical body, so I cannot perform physical tasks or interact with the physical world in any way.*

## 6. Concluding Remarks: Consideration for Ethical and Privacy Concerns

In this paper, we provided a historical overview on the development chatbots. In addition, we looked at the significant technological developments that enabled the emergence and success of ChatGPT. We then described the potential of ChatGPT in several domains and applications. In healthcare, ChatGPT can potentially be used for medical screening, answering general questions, and exam preparation. In education, ChatGPT can be used in tutoring and detecting plagiarism. ChatGPT can also aid researchers with writing, summarizing information, and translating. The emergence of LLMs such as ChatGPT provide several advantages to internet users. Firstly, their ability to understand and generate human-like text leads to improved performance in a multitude of natural language processing tasks, from translation to summarization. Their general-purpose nature contributes to their versatility, making them applicable across various domains such as customer service, content creation, programming help, and education. This broad applicability also means that LLMs can make information and services more accessible, catering to those who prefer or need to use natural language interfaces. Finally, for individual users, LLMs serve as powerful learning and productivity tools. They can assist in learning new topics, brainstorming ideas, and providing writing assistance, effectively becoming an invaluable resource in various personal and professional endeavors. However, there are many ethical and privacy concerns that needs to be addressed about ChatGPT (Zhuo et al., 2023). For instance, some users have reported ChatGPT's responses containing race and gender bias[6]. Moreover, given its effectiveness, ChatGPT may be used for unethical purposes in education, including cheating. In research, ChatGPT raises ethical questions about copyright and plagiarism. Given their training on vast amounts of text data from the internet, LLMs like ChatGPT can generate outputs that resemble or draw heavily from the works they were trained on, even if they do not directly reproduce these texts. For example, if an LLM was trained on a large dataset that includes numerous academic papers, it might generate a new paper that, unintentionally, closely mirrors the ideas, arguments, or specific phrasing found in those source documents. This can raise questions of copyright infringement if the generated content too closely resembles the original copyrighted works. Furthermore, it can lead to issues of plagiarism if the LLM's output is used in a new academic work without appropriate attribution to the original sources that inspired the model's output. In terms of privacy concerns, ChatGPT is trained with more than 300 billion words, potentially containing personal information of internet users[7]. These datasets might inadvertently include personal information from public posts or documents. For instance, if an individual's personal blog post containing their email address or phone number is included in the dataset, the LLM could potentially learn and store that information. Furthermore, as LLMs are designed to learn and improve from user interactions, they could unintentionally process and memorize personal information provided in prompts. For instance, if a user interacts with ChatGPT for a task that involves providing their contact details or other sensitive information, there's a risk that the model could learn and subsequently reproduce this information in future outputs.

## Acknowledgement

The authors would like to thank ChatGPT for providing thoughtful responses to our interview questions.

## Conflicts of Interest

---

The authors declare that they have no conflicts of interest to this work.